\documentclass[conference]{IEEEtran}
\IEEEoverridecommandlockouts
\usepackage{cite}
\usepackage{amsmath,amssymb,amsfonts}
\usepackage{algorithmic}
\usepackage{graphicx}
\usepackage{textcomp}
\usepackage{xcolor}

\usepackage{stfloats}
\usepackage{diagbox}
\usepackage{multirow}
\newcommand{\tabincell}[2]{\begin{tabular}{@{}#1@{}}#2\end{tabular}}

\def\BibTeX{{\rm B\kern-.05em{\sc i\kern-.025em b}\kern-.08em
    T\kern-.1667em\lower.7ex\hbox{E}\kern-.125emX}}
\begin{document}

\title{A Dual-Questioning Attention Network for Emotion-Cause Pair Extraction with Context Awareness\\
\thanks{This work was partially  supported by the Fundamental Research Funds for the Central Universities(DUT20GF106) and the National Natural Science Foundation of China(61806034).}
}

\author{\IEEEauthorblockN{1\textsuperscript{st} Qixuan Sun}
\IEEEauthorblockA{\textit{School of Software Technology} \\
\textit{Dalian University of Technology}\\
Dalian, China \\
i.josh.sun@gmail.com}
\and
\IEEEauthorblockN{2\textsuperscript{nd} Yaqi Yin}
\IEEEauthorblockA{\textit{School of Software Technology} \\
\textit{Dalian University of Technology}\\
Dalian, China \\
yaqiYin@outlook.com}
\and
\IEEEauthorblockN{3\textsuperscript{rd} Hong Yu}
\IEEEauthorblockA{\textit{School of Software Technology} \\
\textit{Dalian University of Technology}\\
Dalian, China \\
hongyu@dlut.edu.cn}
}

\maketitle

\begin{abstract}
Emotion-cause pair extraction (ECPE), an emerging task in sentiment analysis, aims at extracting pairs of emotions and their corresponding causes in documents. This is a more challenging problem than emotion cause extraction (ECE), since it requires no emotion signals which are demonstrated as an important role in the ECE task. Existing work follows a two-stage pipeline which identifies emotions and causes at the first step and pairs them at the second step. However, error propagation across steps and pair combining without contextual information limits the effectiveness. Therefore, we propose a Dual-Questioning Attention Network to alleviate these limitations. Specifically, we question candidate emotions and causes to the context independently through attention networks for a contextual and semantical answer. Also, we explore how weighted loss functions in controlling error propagation between steps. Empirical results show that our method performs better than baselines in terms of multiple evaluation metrics. The source code can be obtained at https://github.com/QixuanSun/DQAN.
\end{abstract}


\section{Introduction}
Emotion cause extraction was firstly defined by ~\cite{Rule-AText-drivenRule-basedSystemforEmotionCauseDetection}, aiming at extracting possible causes in documents with a given emotion. Previously, ECE has aroused a growing number of researches, including rule-based methods ~\cite{Rule-EMOCause,Rule-EmotionCauseDetectionwithLinguisticConstructions,Rule-ExtractingCausesofEmotionsfromText}, machine learning methods ~\cite{ML-EmotionCauseDetectionwithLinguisticConstructioninChineseWeiboText,SVM-Event-DrivenEmotionCauseExtractionwithCorpusConstruction} and neural networks ~\cite{SVM-LSTM-AnEmotionCauseCorpusforChineseMicroblogswithMultiple-UserStructures,Context-awareemotioncauseanalysiswithmulti-attention-basedneuralnetwork,ACo-AttentionNeuralNetworkModelforEmotionCauseAnalysiswithEmotionalContextAwareness,Context-AwareMulti-ViewAttentionNetworksforEmotionCauseExtraction,MultipleLevelHierarchicalNetwork-BasedClauseSelectionforEmotionCauseExtraction,ARNN-TransformerHierarchicalNetworkforEmotionCauseExtraction,JointLearningforEmotionClassificationandEmotionCauseDetection,ExtractingEmotionCausesUsingLearningtoRankMethodsFromanInformationRetrievalPerspective,AdversarialTrainingbasedCross-lingualEmotionCauseExtraction,AKnowledgeRegularizedHierarchicalApproachforEmotionCauseAnalysis}. However, such approaches have a high dependency on labelled emotions. ~\cite{Emotion-CausePairExtraction:ANewTasktoEmotionAnalysisinTexts} experimented with a removal of annotated emotions on ECE methods, and the F1 score drops dramatically about 34.69\% \cite{ACo-AttentionNeuralNetworkModelforEmotionCauseAnalysiswithEmotionalContextAwareness}.

\begin{figure}[t]
\centering
\includegraphics[width=\columnwidth]{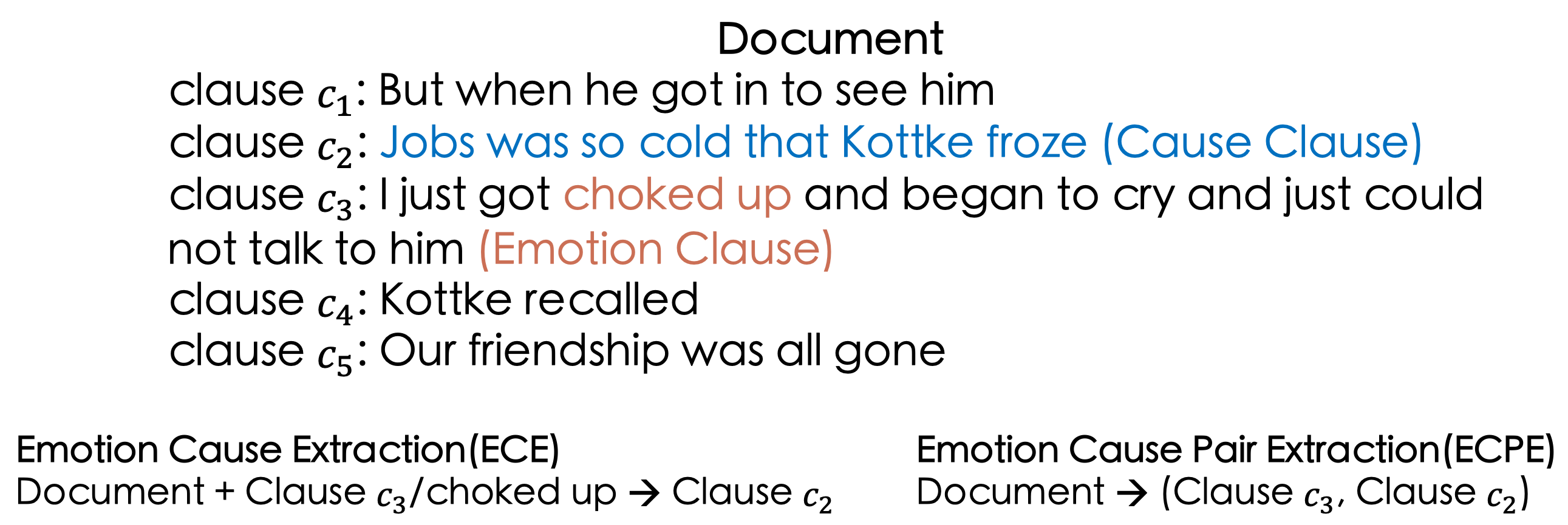}
\caption{An intuitive example of differences between ECE and ECPE task.}
\label{diff}
\end{figure}

Expensive labelling limit the application of ECE-driven techniques. To address the limitation, recent works focused on the emotion-cause pair extraction (ECPE) task \cite{Emotion-CausePairExtraction:ANewTasktoEmotionAnalysisinTexts}, which aims to extract pairs of emotions and their corresponding causes in a document. Figure~\ref{diff} is an intuitive example of differences between ECE and ECPE task. Compared with ECE task, ECPE task is more challenging as it requires no transcendental information of annotated emotions and extracts emotions, causes and emotion-cause pairs from plain text. Existing approaches follow a two-step pipeline \cite{Emotion-CausePairExtraction:ANewTasktoEmotionAnalysisinTexts}: detecting emotions and causes firstly and then pairing them from a Cartesian product of emotions and causes. However, these methods overlook the error propagation from the first stage to the second. Intuitively, when an emotion or cause is predicted as negative in step one, it will not participate in step two and completely becomes negative. Table~\ref{study1} presents the recall rate of two successive steps. The recall rate of the second step is highly correlated to that of the first step. In addition, \cite{Emotion-CausePairExtraction:ANewTasktoEmotionAnalysisinTexts} ignores contextual and semantic information when pairing emotions and causes, which is essential for ECPE. For example, a student would probably be happy when he heard he got an A in a class, but would also possibly be sad if his goal was to pass all the courses with an A+. In this way, a common emotion-cause pair can be a wrong pair given a particular context. 

\begin{table}[h]
\caption{Results of recall rate of three sub-tasks in ECPE task generated by three models \cite{Emotion-CausePairExtraction:ANewTasktoEmotionAnalysisinTexts}. More details are given in Section~\ref{Baselines}}
\centering
\begin{tabular}{l|cc|c}
\hline
 & \multicolumn{2}{c|}{Step1} & Step2 \\ \hline
 & \tabincell{c}{Emotion\\ Extraction}  & \tabincell{c}{Cause\\ Extraction}  & \tabincell{c}{Emotion-cause\\ Pair Extraction}\\ \hline
Indep & 0.8071 & 0.5673  & 0.5082 \\
Inter-CE & 0.8122 & 0.5634  & 0.5135 \\ 
Inter-EC & 0.8107 & 0.6083 & 0.5705 \\
\hline
\end{tabular}
\label{study1}
\end{table}

In order to tackle these problems, we propose a model called Dual-Questioning Attention Network (DQAN). Specifically, weighted loss function is applied to address error propagation and better focus on imbalanced text. We introduce a hierarchical network with two levels including word level and clause level. A novel dual-questioning mechanism is applied to jointly consider the representation of candidate pairs of emotion and cause and the rest of the context. Features of candidate emotions and causes are extracted and matched through the contextual information for fully utilization. Furthermore, we propose a method to incorporate and encode distance information in pair extraction. To evaluate the effectiveness of our model, we conduct experiments on benchmark ECPE dataset and ECE datasets. The main contributions of our work can be summarized as follows:
\begin{itemize}
\item We propose a dual-questioning attention network to handle the ECPE task in a unified context, which learns useful feature for pairing and matching across the semantics.
\item We employ a weighted loss function for the imbalanced task, which alleviate and moderate error propagation appeared in previous works.
\item We conduct experiments on benchmark datasets to show the effectiveness of our model and the performance shows that our model significantly outperforms the baseline models.
\end{itemize}

The rest of this paper is organized as follows. Section~\ref{Related work} presents related work in emotion cause pair extraction and Section~\ref{Model} introduces our model. Experiments set up and results are reported in Section~\ref{Experiments}, followed by conclusions in Section~\ref{Conclusions}.

\section{Related work}\label{Related work}
Lee et al.~\cite{Rule-AText-drivenRule-basedSystemforEmotionCauseDetection} firstly restrict the definition of automatic emotion cause extraction. They develop two sets of rules for emotion cause detection and propose an evaluation scheme. Rule based methods have been improved by later researchers. Russo et al.\cite{Rule-EMOCause} proposed a rule based on relevant linguistic patterns and an incremental repository of common sense knowledge on emotional states. Chen et al.\cite{Rule-EmotionCauseDetectionwithLinguisticConstructions} created two sets of linguistic patterns. Syntactic and dependency parser and rules were added up to the principle by~\cite{Rule-ExtractingCausesofEmotionsfromText}. Other than rule based methods, there were also some machine learning based models. Gui et al.\cite{ML-EmotionCauseDetectionwithLinguisticConstructioninChineseWeiboText} employed SVMs and CRFs on extraction causes. Gui et al.\cite{SVM-Event-DrivenEmotionCauseExtractionwithCorpusConstruction} applied multi-kernel SVM on emotion-cause identification. They also presented an open-source dataset using SINA city news, which led to next period of prospectives in emotion cause extraction task.

Deep learning networks have been proved efficient on other NLP tasks such as sentence classification and name entity identification. Therefore, a variety of neural networks have been applied to the task. One approach to emotion cause detection is to treat it as clause classification problem. With the huge success of hierarchical attention network (HAN) \cite{HAN} on document classification, hierarchical models have been widely used on various kinds of NLP task. Li et al.\cite{Context-awareemotioncauseanalysiswithmulti-attention-basedneuralnetwork} and Li et al.\cite{ACo-AttentionNeuralNetworkModelforEmotionCauseAnalysiswithEmotionalContextAwareness} both proposed a hierarchical model, which encoded clauses with bi-directional long short term memory and applied attention mechanism to the emotion clause and applicant cause clauses. However, they simply input only one applicant cause clause at one time, which ignored information in the rest of the article and relationships between them. Other approach is to apply sequence model to extract cause clause. Cheng et al.\cite{SVM-LSTM-AnEmotionCauseCorpusforChineseMicroblogswithMultiple-UserStructures} proposed a basic long short-term memory to the task. Gui et al.\cite{AQuestionAnsweringApproachtoEmotionCauseExtraction} presented a convolutional Multiple-Slot deep memory network. Other researchers mostly presented a hierarchical model, that bi-directional long short-terms memory was often applied on encoding clauses and differences appeared in clause-level network. Xiao et al.\cite{Context-AwareMulti-ViewAttentionNetworksforEmotionCauseExtraction} proposed a multi-view attention network (COMV) and Yu et al.\cite{MultipleLevelHierarchicalNetwork-BasedClauseSelectionforEmotionCauseExtraction} applied attention mechanism on word-level and convolutional network on phrase-level. Xia et al.\cite{ARNN-TransformerHierarchicalNetworkforEmotionCauseExtraction} presented a joint emotion cause extraction framework, called RNN-Transformer Hierarchical Network (RTHN), to efficiently capture causality between clauses.

Xia et al.\cite{Emotion-CausePairExtraction:ANewTasktoEmotionAnalysisinTexts} pointed out the limitation of previous research on ECE and defined the task of emotion-cause pair extraction (ECPE). In their approaches, pair extraction follows a pipeline framework, including extracting candidate emotion and cause clauses and classifying the relations between them subsequently. They proposed three different models, which extracting both clauses independently or enhancing one's extraction through another. However, they ignored error propagation, which is a commonplace in pipeline models. Besides, contextual information is also ignored in their work \cite{Emotion-CausePairExtraction:ANewTasktoEmotionAnalysisinTexts}, which is necessary for pair extraction. Our model is proposed to address the mentioned problems. To be more specific, our model learns the contextual information through a novel structure, namely dual-questioning attention and shares the contextual information with candidate emotions and causes. In addition, we employ a weighted loss function to alleviate error propagation and develop a method to joint distance information with the whole context.

\section{Methodology}\label{Model}

This section introduces the overall and specific architecture of our model. Our method consists of two stages: extracting possible emotions and causes, and then pairing them.

\subsection{Task definition}
Suppose we have a document $D$, containing several clauses as $d=\{c_1,...,c_i,...,c_{|d|}\}$, where $c_i$ is the $i$-th clause in $d$. Each clause $c_i$ consists of $|c_i|$ words, as $c_i = \{w_{i}^1,...w_{i}^t,...w_{i}^{|c_i|}\}$. To prevent confusion, we apply a superscript 'e' on true emotion clause, and 'c' on true cause clause. In this work, our goal is to identify whether a clause pair $(c_i, c_j) (i, j=1,2,...|d|)$ is the true emotion-cause pair $(c^e, c^c)$ in the passage. Therefore, we can form ECPE task as a binary classification task:
\begin{equation}
f(c_i, c_j) = 
\left\{
\begin{aligned}
& 1;\quad  if \; c_i  = c^c \; and \; c_j = c^e \\
& 0;\quad  otherwise
\end{aligned}
\right.
\end{equation}

\subsection{Overall architecture}
\begin{figure*}[ht]
\centering
\includegraphics[width=\textwidth]{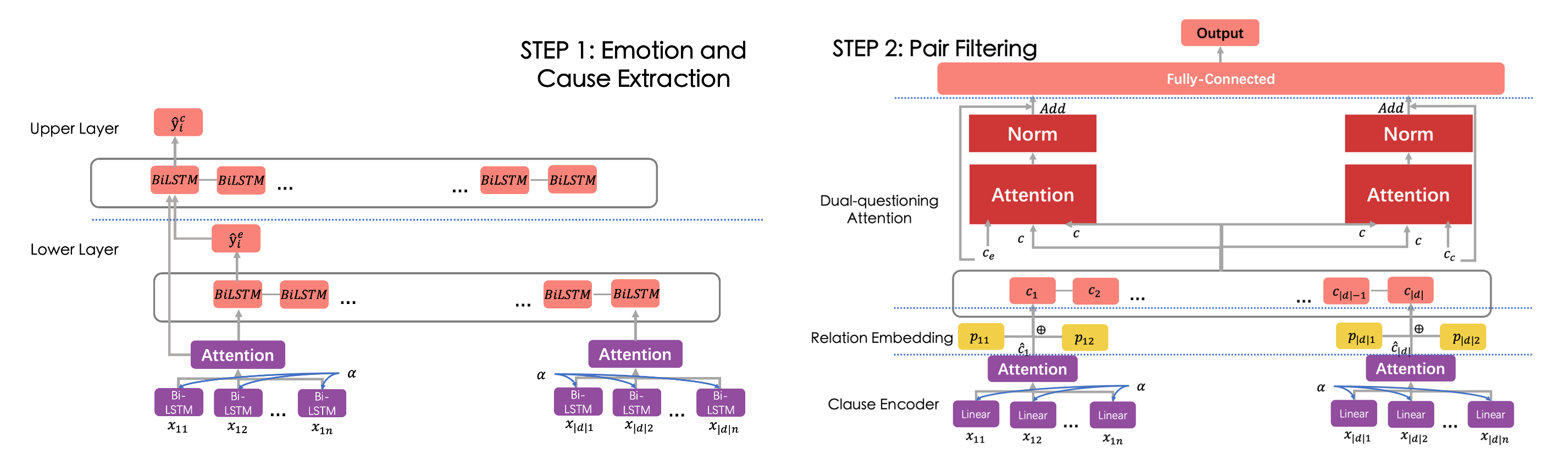}
\caption{Model architecture}
\label{model}
\end{figure*}

The overall architecture of our work is shown in Figure~\ref{model}, which consists of two steps. The first step aims to extract a set of emotion clauses and a set of cause clauses in each document respectively and the second step yields true pairs based on candidate emotion and cause clauses extracted from step one and semantic contexts.

\subsection{Step1: Emotion and cause extraction}
In step one, we apply a dual-layer sequential network to jointly extract emotion and cause clauses. Considered the relationship between emotions and its causes, we extract emotion clauses on the lower layer firstly and cause clauses on the upper layer subsequently. Given a clause $c_i = \{w_{i}^1,...w_{i}^t,...w_{i}^{|c_i|}\}$, we firstly embeds words with embedding matrix $W_e$ to get an input sequence word embedding $x_{i}^t$. Then a bi-directional LSTM is applied to yield contextual representation of a word and an attention mechanism is utilized to generate a clause representation $c_i$.
\begin{align}
& h_i^t = {\rm BiLSTM}(x_i^t) \\
& {\alpha}_{i}^t =\dfrac{{\rm exp}({h_{i}^t}^Tu_w)}{\sum_t {\rm exp}({h_{i}^t}^Tu_w)} \\
& c_i = \sum_t {\alpha}_{i}^th_{i}^t
\end{align}
where $u_w$ is a learnable weight parameter.

Secondly, we construct a dual-layer network with bi-LSTMs, linear projections and softmax layer for emotion and cause extraction based on clause representations $c_i$. In addition, considered the enhancement of cause extraction with emotion extracted, we firstly detect emotion clauses and then after extract cause clauses.
\begin{align}
& \hat{y_i^e} = {\rm softmax}(W_c {\rm BiLSTM}(c_i)) \\
& \hat{y_i^c} = {\rm softmax}(W_e {\rm BiLSTM}(c_i \oplus \hat{y_i^e}))
\end{align}
where $W_c, W_e$ represents projection matrices.

Since extraction task is heavily imbalanced, we apply a weighted loss with two components for training procedure.
\begin{equation}
L =  \sum_{i}\beta_1(y_{i}^c \log(\hat{y_i}^{c}) + y_{i}^e \log(\hat{y_i}^{e}))
\end{equation}
where $y_i^c, y_i^e$ is the ground-truth of $\hat{y}_i^c, \hat{y}_i^e$, and $\beta_1$ is the weight vector of weighted loss.

\subsection{Step 2: Pair matching}

\paragraph{\bf Clause encoder} 
Word-level encoder aims to obtain a clause representation $\hat{c}_i$ from sequence word vectors.
Given a clause $c_i = \{w_{i}^1,...w_{i}^t,...w_{i}^{|c_i|}\}$, we first embed the words with embedding matrix $W_e$ to get an input sequence word embedding $x_{i}^t$ and its hidden representation $h_{i}^t$ with a linear projection.
\begin{equation}
h_{i}^t = W_hx_{i}^t + b_h
\end{equation}

When generating clause representations, not all text words make the same contribution. The attention mechanism produces a context vector by focusing on different portions of the word sequence and aggregating the hidden representations of those representative words. Specifically, the weight ${\alpha}_{it}$ is computed as follows with the attention mechanism:
\begin{align}
u_{i}^t &= {\rm tanh}\left(W_wh_{i}^t+b_w\right) \\
{\alpha}_{i}^t &=\dfrac{{\rm exp}({u_{i}^t}^Tu_w)}{\sum_t {\rm exp}({u_{i}^t}^Tu_w)} 
\end{align}

where $W_w, b_w, u_w$ are weight parameters and $h_{i}^t$ is the current hidden representation of the $t$-th word in clause $c_i$. Then, the weight ${\alpha}_{i}^t$ is combined with hidden representations $h_{i}^t$ to get the clause representation $\hat{c}_i $ as follows:
\begin{equation}
\hat{c}_i = \sum_t {\alpha}_{i}^th_{i}^t
\end{equation}

\paragraph{\bf Relation embedding}
Relation embedding was often applied to describe the relative position of a word from two named entities in relation extraction tasks \cite{RelationExtractionPCNN} and performed well. In our work, clause position embedding is utilized to compute relative distance from current clause to cause clause and emotion clause respectively. The clause representation can be computed by following:
\begin{equation}
c_i = \hat{c}_i \oplus p_{i1}\oplus p_{i2}
\end{equation}
$\hat{c}_i$ refers to raw clause representation from output of word-level layer and $p_{i1}$, $p_{i2}$ are relative positions based on cause clause and emotion clause.

\paragraph{\bf Dual-questioning attention}

Inspired by the outstanding performance in sentence encoding by ~\cite{Transformer}, we propose a dual-questioning attention to better focus on semantic understanding and pair extracting in ECPE task. Firstly, a simple attention function can be described as mapping a query and a set of key-value pairs to an output:

\begin{equation}
{\rm Att}(Q, K, V) = {\rm softmax}(\dfrac{QK^T}{\sqrt{d_k}})V
\end{equation}

where matrice $Q, K, V$ refer to the query, keys and value, and $d_k$ is the dimension of query and keys. 

Then, we propose a dual-questioning attention since we question the candidate emotion and cause to the rest of the context respectively, each of which can be produced with a linear projection of several outputs of simple attention mechanism. 

\begin{align}
&
\begin{aligned}
\hat{z_1} = & [{\rm Att}(c_e W_{11}^{c_e}, c W_{11}^{c}, c W_{11}^{c})  \\
&\oplus ... \oplus {\rm Att}(c_e W_{i1}^{c_e}, c W_{i1}^{c}, c W_{i1}^{c})]W^O \\
\end{aligned}
\\
& 
\begin{aligned}
\hat{z_2} = & [{\rm Att}(c_c W_{12}^{c_c}, c W_{12}^{c}, c W_{12}^{c})  \\
& \oplus ... \oplus {\rm Att}(c_c W_{i2}^{c_c}, c W_{i2}^{c}, c W_{i2}^{c})]W^O \\
\end{aligned}
\end{align}

where $W_{ij}^{c} \in \mathbb{R}^{d_{c} \times d_c}$, $W_{ij}^{c_e} \in \mathbb{R}^{d_{c_e} \times d_{c_e}}$ and $W_{ij}^{c_c} \in \mathbb{R}^{d_{c_c} \times d_{c_c}}$  are projection matrices and $d_{c}, d_{c_e}, d_{c}$ refer to the dimension of the representation of a sequence of clause $c = \{c_1, c_2, ..., c_{|d|}\}$, cause clause $c_c$ and emotion clause $c_e$. The $i$ represents the number of heads that we generate outputs from the simple attention mechanism, which is a hyper-parameter that we set in the next section.

Then, the output $\hat{z_1}, \hat{z_2}$ are combined with the query of inputs and fed into a normalization layer as follows.
\begin{equation}
z = {\rm Norm}(\hat{z_1}\oplus c_e) \oplus {\rm Norm}(\hat{z_2}\oplus c_c) 
\end{equation}

\paragraph{\bf Classification}
The fully connected layer is used on classification according to features produced by the normalization layer. The output is then computed with a linear transformation by following:
\begin{equation}
o = wz+b
\end{equation}
$w$ and $b$ are weighted matrix and bias vector, and output vector $o \in \mathbb{R}^2$ is the output of fully connected layer. We finally add a softmax function to compute probability distribution $\hat{y}_i$ from the output vector.
\begin{equation}
\hat{y}_i = \dfrac{\exp(o_i)}{\sum_i \exp(o_i)}
\end{equation}

\paragraph{\bf Training procedure}
For pair filtering, we have more negative candidate pairs which must lead to a more imbalanced problem. In this way, we apply a weighted loss function to address the issue, focusing on positive pairs.
\begin{equation}
L_{pair} = \sum_{c_e, c_c}(\beta_2 y_{c_e,c_c} \log(\hat{y}_{c_e,c_c})) + \lambda||\theta||^2
\end{equation}
where $y_{c_e,c_c}$ is the ground-truth of $\hat{y}_{c_e,c_c}$, and $\beta_2$ is the weight vector of weighted loss. A $l2$-normalization is also added to the loss function with $\lambda$ restricted.

\section{Experiments}\label{Experiments}

\subsection{Datasets and experimental settings}
We conducted experiments on benchmark ECPE dataset released by \cite{Emotion-CausePairExtraction:ANewTasktoEmotionAnalysisinTexts}. Table \ref{ECPE dataset} presents details of the dataset.  The
metrics we used in evaluation follows\cite{Emotion-CausePairExtraction:ANewTasktoEmotionAnalysisinTexts}. We use precision, recall and f1-score as evaluation metrics, which are calculated as follows:
\begin{equation}
\begin{aligned}
&P = \dfrac{\sum_{correct\_pairs}}{\sum_{proposed\_pairs}} \\
&R = \dfrac{\sum_{correct\_pairs}}{\sum_{annotated\_pairs}} \\
&F1 = \dfrac{2\times P \times R}{P+R} \\
\end{aligned}
\end{equation}
where proposed pairs denotes the emotion cause pairs predicted by the model, annotated pairs denotes the emotion cause pairs that are labeled in the dataset and correct pairs means the pairs that are both labeled and predicted as an emotion cause pair.

\begin{table}[h]
\centering
\caption{Details of the ECPE dataset.}
\label{ECPE dataset}
\begin{tabular}{l|c|c} \hline
 & Number & Percentage \\ \hline
Documents & 1945 & 100\% \\
Documents with 1 emotion-cause pair & 1746 & 89.77\% \\
Documents with 2 emotion-cause pairs & 177 & 9.10\% \\
Documents with $\ge$ 2 emotion-cause pairs & 22 & 1.13\% \\ \hline
\end{tabular}
\end{table}

The word vectors \cite{word2vec} are pre-trained on the corpora from Chinese Weibo by word2vec and keep unchanged during the training duration. The dimension of word embedding and relative position embedding are set to be 200 and 50 respectively. The number of heads in multi-questioning attention is set to 4. In order to avoid overfitting, we apply dropout to embeddings and inputs of softmax layer, which is set to 0.5 and 0.8. Besides, L2-normalization is set as 1e-5. For training details, we use the Adam optimization with shuffled batches. Batch size and learning rate are set to 128 and 0.005. The best weight parameters in stage one and two are 3 and 2.5, which will be discussed in next section. Furthermore, all weight matrices and bias are randomly initialized by a uniform distribution U (-0.01, 0.01).

\subsection{Baselines}\label{Baselines}
We compare our approach with the following baselines:
\begin{itemize}
\item {\bf Indep} independently extracts emotions and causes by two Bi-LSTMs and then pairs the extracted emotions and causes with a simple classifier \cite{Emotion-CausePairExtraction:ANewTasktoEmotionAnalysisinTexts}.
\item {\bf Inter-CE} generally follows the procedure of Indep, however, enhances emotion extraction with the predictions of cause extraction \cite{Emotion-CausePairExtraction:ANewTasktoEmotionAnalysisinTexts}.
\item {\bf Inter-EC} generally follows the procedure of Indep, however, enhances cause extraction with the predictions of emotion extraction \cite{Emotion-CausePairExtraction:ANewTasktoEmotionAnalysisinTexts}.
\item {\bf Indep-W} is an extended model of {\bf Indep} with weighted loss on two stages.
\item {\bf Inter-CE-W}  is an extended model of {\bf Inter-CE} with weighted loss on two stages.
\item {\bf Inter-EC-W}  is an extended model of {\bf Inter-EC} with weighted loss on two stages.
\end{itemize}

\subsection{Results}
For the purpose of simplicity, we denote the proposed model as DQAN. We report related evaluation metrics of our method and all baselines on the test sets.
\begin{table}[h]
\centering
\caption{Experiment results of all proposed models and baselines in emotion-cause pair extraction with precision, recall, and F1-measure as metrics.}
\label{ECPE}
\begin{tabular}{l|ccc}\hline
 &  \multicolumn{3}{c}{Emotion-cause Pair Extraction} \\ \hline
Methods & Precision & Recall & F1 \\ \hline
{\bf Indep \cite{Emotion-CausePairExtraction:ANewTasktoEmotionAnalysisinTexts}} & 	0.6832&	0.5082	&	0.5818\\
{\bf Inter-CE  \cite{Emotion-CausePairExtraction:ANewTasktoEmotionAnalysisinTexts}} &	{\bf 0.6902}&	0.5135	&	0.5901\\
{\bf Inter-EC  \cite{Emotion-CausePairExtraction:ANewTasktoEmotionAnalysisinTexts}} &	0.6721&	0.5705	&	0.6128\\  \hline
{\bf Indep-W} & 	0.6155	&0.5768	&0.5940\\
{\bf Inter-CE-W} 	&0.6206	&0.5910	&0.6047\\
{\bf Inter-EC-W} & 0.6658	&0.5823	&0.6205\\ \hline
{\bf DQAN}&    0.6733& {\bf 0.6040}   &   {\bf  0.6362}\\ \hline
\end{tabular}
\end{table}

The experimental results of our method and the baselines on ECPE dataset are shown in Table~\ref{ECPE}. Our results are performed with $p<0.05$ tested by the Student’s paired t-test. Compared with {\bf Inter-EC}, {\bf DQAN} shows an improvement of 3.82\% f1-score, 5.87\% recall and 0.16\% precision in emotion-cause pair extraction. It indicates that weighted loss works well in reduction of error propagation and attention mechanism properly integrates contextual semantics. 

Furthermore, in order to confirm the effectiveness of our weighted loss in alleviating error propagation by enhancing recall rate of the two steps presented in Table \ref{study1}, we experiment existing methods with our weighted loss. The approaches {\bf Inter-EC-W, Inter-CE-W, Indep-W} are extended models of {\bf Inter-EC, Inter-CE, Indep} with weighted loss on two stages. For the f1 measurement, the three derivative models achieve 1.26\%, 2.47\% and 2.10\% higher than their origin models respectively, which proves that our weighted loss is beneficial on ECPE task. Beyond that, our {\bf DQAN} outperforms 3.73\% and 2.53\% on recall rate and f1-score compared with {\bf Inter-EC-W}. The difference between {\bf Inter-EC-W} and {\bf DQAN} is that {\bf DQAN} uses a dual-questioning attention module to obtain the emotion and cause representation from context while {\bf Inter-EC-W} ignores context and simply classifies emotions and causes. The better performance on recall rate and f1-score indicates that dual-questioning attention module can effectively combine with weighted loss to push it further on alleviation of error propagation by  jointly considering contextual semantics.

\subsection{Discussion}

\subsubsection{Ablation study}
To understand the efficacy of dual-questioning attention modules and weighted loss functions, we conduct ablation experiment on these modules.

\begin{table}[h]
\centering
\caption{Ablation study for the DQAN model}
\label{ablation}
\begin{tabular}{l|c|c|c} \hline
Methods & P & R & F1 \\ \hline
DQAN & 0.6733& 0.6040   &  {\bf  0.6362}\\ 
DQAN w/o dual-questioning &0.6728&0.5815 &0.6234 \\ 
DQAN w/o weighted loss 1 &{\bf 0.7089}	&0.5503	&0.6184 \\
DQAN w/o weighted loss 2 &0.6418	&{\bf 0.6082}	&0.6234 \\
Inter-EC-W &0.6658	&0.5823	&0.6205\\ \hline
\end{tabular}
\end{table}

From Table~\ref{ablation}, we can observe that after reducing dual-questioning attention, the recall and f1-score of {\bf DQAN} drop 3.59\% and 2.47\%. In addition, {\bf DQAN} without dual-questioning attention module performs practically as same as {\bf Inter-EC-W}, which setups the same weight and ignores contextual semantics. The main reason is that dual-questioning attention is capable, enforcing the model to pay more attention on contextual semantics that can enhance the connection between emotions and causes. Besides, with a removal of weighted loss function in step one and two, the f1-score drop about 2.80\% and 2.01\% respectively, which proves our weighted loss function is effective in capturing emotion-cause pairs. Between such two weighed loss functions, weighted loss in step one is more essential on cutting error propagation. The precision is the highest when removing weighted loss one and the recall rate is the highest when removing weight loss two, mainly because the former aims to enhance and expand candidate emotion-cause pairs while the latter aims to narrow then down.

\subsubsection{Effects of weighted loss}

During the experiment, we find that the two weighted loss contributes significantly to the results. In order to explore the real influence of weighted loss, we set different weight on weighted loss and leave other parameters as they are, and observe evaluation on ECPE task.

\begin{figure}[h]
\centering
\includegraphics[width=0.49\linewidth]{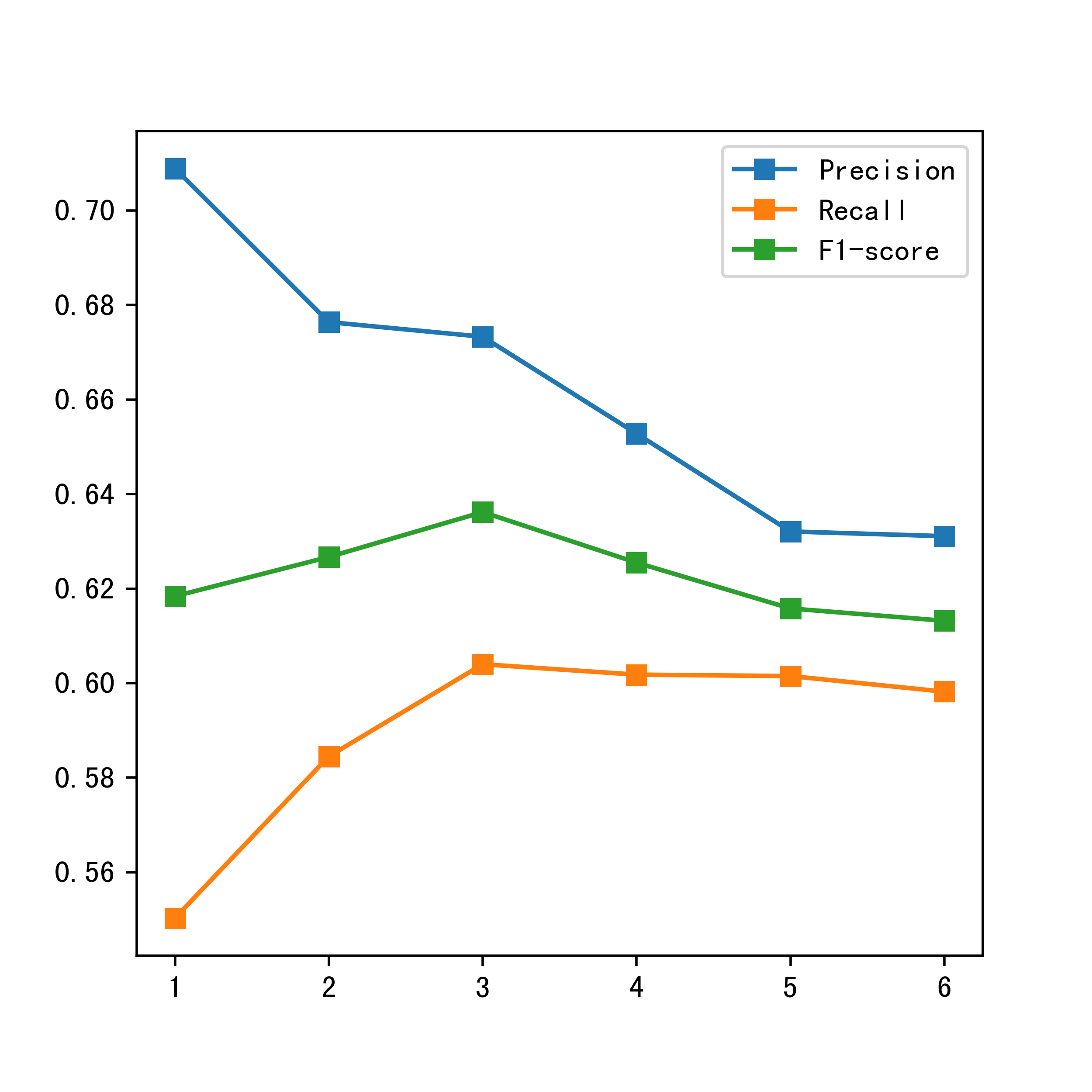}
\includegraphics[width=0.49\linewidth]{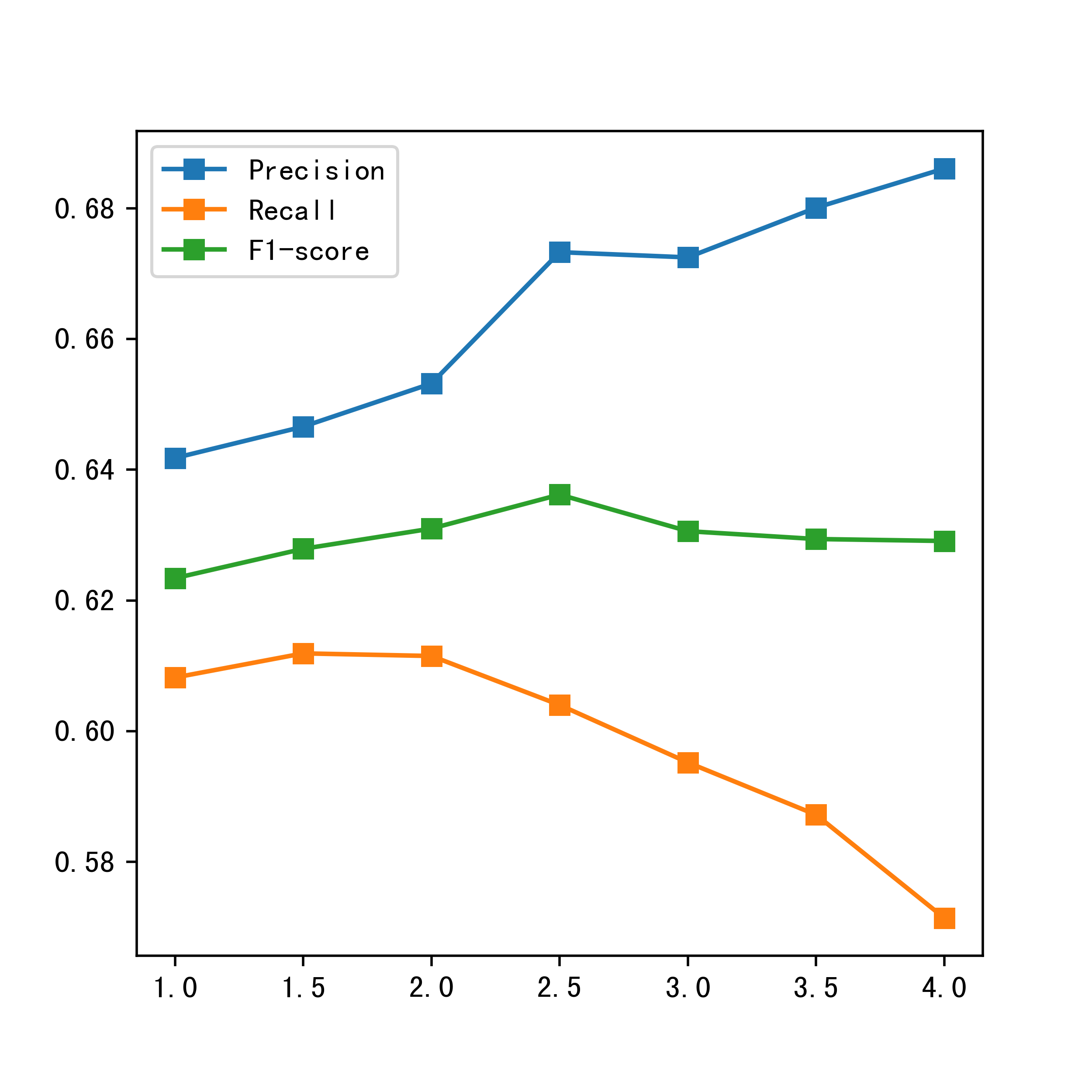}
\caption{Weight variation of weighted loss in stage one and two\label{subfig:left}}
\end{figure}

Figure~\ref{subfig:left} firstly illustrates how the performance varies on different weight parameters in stage one. With the increase of the value of weight, we can see an increase of recall and a decrease of precision, and that both a small weight and a large weight will cause f1-score to drop. This is because a greater weight on positive clauses could enlarge the recall rate of emotion and cause extraction in stage one and lead to an increase of recall of emotion-cause pair extraction. But it's also a burdensome to the model of stage two, which result in a decrease of precision. Therefore, the highest f1-score appears when a proper weight is set because f1-score is a harmonic mean of precision and recall. It also shows a performance variation with varied weight  parameters in stage two, a practically mirrored picture of that in stage one. This is because we set a greater weight on negative pairs in stage two instead of positive labels in stage one. Both of the results show that a proper weight is essential for ECPE task.

\subsubsection{Runtime analysis}

\begin{figure}[h]
\centering
\includegraphics[width=0.49\linewidth]{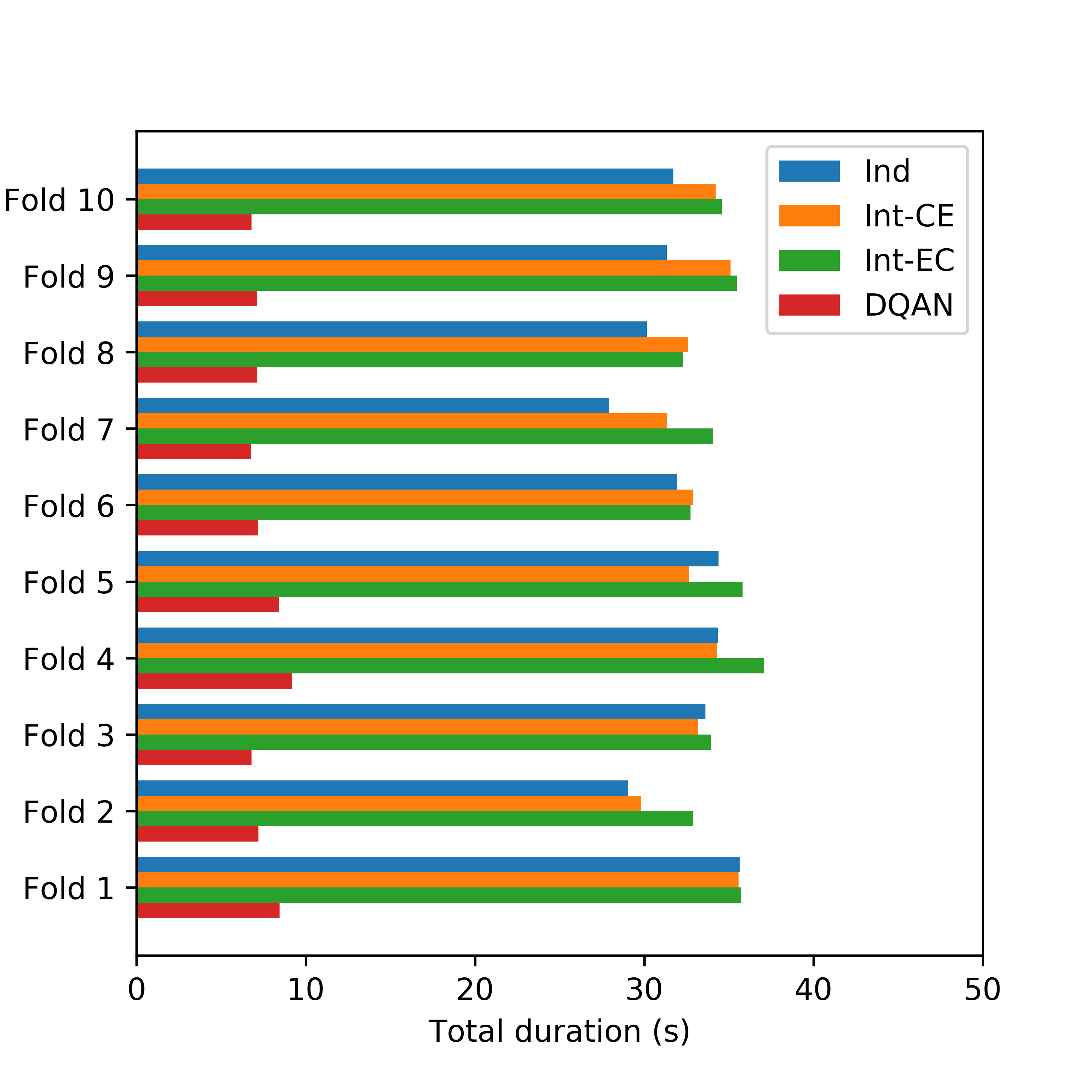} 
\includegraphics[width=0.49\linewidth]{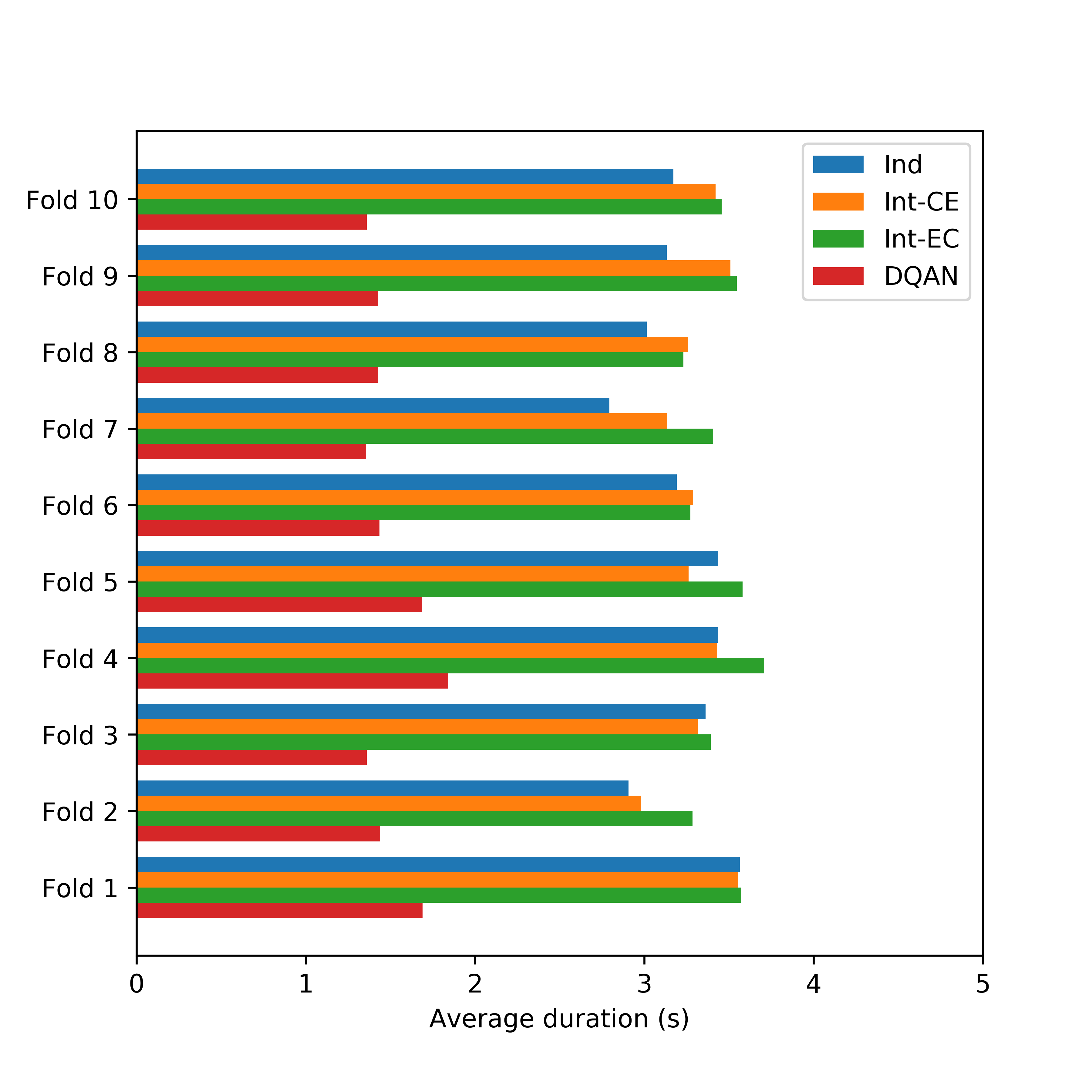}
\caption{Runtime on each fold of overall epochs and per epoch\label{subfig:right}}
\end{figure}

To confirm that our method is more efficient than baselines, we perform runtime analysis among them. Note that our method needs only 5 train epochs on each fold, while baselines need 10 epochs. Therefore, we report the average running time on one epoch and total running time on all epochs on each fold. Results showed in Figure~\ref{subfig:right} suggests that our method reduce a 75\% and 55\% of time on total and average runtime on step two compared with previous methods \cite{Emotion-CausePairExtraction:ANewTasktoEmotionAnalysisinTexts}.

\subsubsection{Error analysis}

We are also curious about the reason of existing errors, therefore, we conduct further analysis on wrong pairs extracting in step two, traceback their origins, and classify errors into four categories. Table~\ref{error} shows detailed error analysis of two models, DQAN and Inter-EC\cite{Emotion-CausePairExtraction:ANewTasktoEmotionAnalysisinTexts}. In this section, we mostly talk about False Negative samples (FN), which is also the major reason for error propagation as we talked in previous sections. Given a false negative sample $(c_e, c_c)$, there are four types of reasons, which can be divided into the fault of step one (3) and step two (1). The fault of step one represents the deficiency of parts of correct pairs sending into step two, including no emotion clause $(None, c_c)$ (no emo.), no cause clause $(c_e, None)$ (no cau.), neither emotion clause nor cause clause $(None, None)$ (no emo. and cau.). The fault of step two is wrong pairing that $f(c_e, c_c)=0$. We present the average number of errors per fold (avg.) in Table ~\ref{error}. The fault of step one account for 99.16\% of all errors in {\bf Inter-EC}, which proves that the first three types of errors are main issues of all wrong cases. Our DQAN model drops about 17.00\%, 12.45\% and 18.67\% on these three types of errors compared with {\bf Inter-EC} respectively, indicating that our model can alleviate these errors significantly. The decrease testifies the effectiveness of weighted loss to resist from error propagation. 

\begin{table}[ht]
\centering
\caption{The average number and percentage of errors per fold.}
\label{error}
\begin{tabular}{c|c|c|c|c|c}
\hline
\diagbox [width=5em,trim=l] {Methods}{type} & no emo. & no cau. &  \tabincell{c}{no emo.\\ and cau.} & wrong pair & Total \\ \hline
\multirow{2}*{\bf Inter-EC} & 10.0 & 53.0 & 31.6 & 0.8 & 95.4 \\
& 10.48\% & 55.56\% & 33.12\% & 0.84\% & 100\% \\ \hline
\multirow{2}*{\bf DQAN} & 8.3 & 46.4 & 25.7 & 5.8 & 86.2 \\
& 9.63\%  & 53.83\% & 29.81\% & 6.73\% & 100\% \\ \hline
\end{tabular}
\end{table}

\subsection{Further results on ECE task}
\begin{table*}[h]
\centering
\caption{Experiments results on Chinese and English datasets on ECE task. Methods are divided into ECE based and ECPE based,  and the bold represents the highest performance on each type of methods. }\label{ECE}
\begin{tabular}{l|ccc|ccc} \hline
 & \multicolumn{3}{c|}{Chinese dataset} & \multicolumn{3}{c}{English dataset}  \\ \hline
Methods & P & R & F1 & P & R & F1 \\ \hline
 \emph{ECE methods}  & \multicolumn{6}{c}{} \\ \hline
 {\bf Word2vec \cite{word2vec}} & 0.4301 & 0.4233 & 0.4136 & 0.1651 & {\bf 0.8673} & 0.2774\\
{\bf SVM} & 0.4200 & 0.4375 & 0.4285 & 0.2757 & 0.6416 & 0.3856\\
{\bf CNN \cite{CNN}} & 0.6472 & 0.5493 & 0.5915 & 0.7218 & 0.2628 & 0.3390\\
{\bf Multi-Kernel \cite{SVM-Event-DrivenEmotionCauseExtractionwithCorpusConstruction}}& 0.6588 &0.6927 &0.6752 & - & - & -\\
{\bf Memnet \cite{AQuestionAnsweringApproachtoEmotionCauseExtraction}}& 0.5922 &0.6354 &0.6134 & - & - & -\\
{\bf ConvMS-M \cite{AQuestionAnsweringApproachtoEmotionCauseExtraction}} &0.7076 &0.6838 &0.6955 & 0.4605 &0.4177 & 0.4381\\
{\bf CANN \cite{ACo-AttentionNeuralNetworkModelforEmotionCauseAnalysiswithEmotionalContextAwareness}} &0.7721 &0.6891 &0.7266 & - & - & -\\ 
{\bf RTHN \cite{ARNN-TransformerHierarchicalNetworkforEmotionCauseExtraction}} &0.7697 & {\bf 0.7662} &0.7677 & - & - & -\\  
{\bf MANN \cite{Context-AwareMulti-ViewAttentionNetworksforEmotionCauseExtraction}} & {\bf 0.7843} & 0.7587 & {\bf 0.7706} & {\bf 0.7933} & 0.4081 & {\bf 0.5328}\\
\hline
 \emph{ECPE methods} & \multicolumn{6}{c}{} \\ \hline
{\bf CANN-E \cite{Emotion-CausePairExtraction:ANewTasktoEmotionAnalysisinTexts}} &0.4826 &0.3160 &0.3797 & - & - & -\\
{\bf Inter-EC \cite{Emotion-CausePairExtraction:ANewTasktoEmotionAnalysisinTexts}}& 0.7041 &0.6083 &0.6507 & - & - & -\\ \hline
{\bf DQAN} &{\bf 0.7732} &{\bf 0.6370} & {\bf 0.6979}  &{\bf  0.8058} & {\bf 0.4324} & {\bf 0.5628}\\ \hline
\end{tabular}
\end{table*}

Since previous works conducted experiments on a Chinese ECE dataset\cite{SVM-Event-DrivenEmotionCauseExtractionwithCorpusConstruction} with their ECPE methods, we evaluate our model's performance on ECE tasks as well. Furthermore, we expand our experiments on another English dataset\cite{NTCIR}, which is also a benchmark dataset on ECE task. The details about the two datasets are summarized in Table \ref{ECE dataset}. The
metrics we used in evaluation follows\cite{Rule-AText-drivenRule-basedSystemforEmotionCauseDetection} with precision, recall and f1-score used as evaluation metrics.
\begin{equation}
\begin{aligned}
&P = \dfrac{\sum_{correct\_causes}}{\sum_{proposed\_causes}} \\
&R = \dfrac{\sum_{correct\_causes}}{\sum_{annotated\_causes}} \\
&F1 = \dfrac{2\times P \times R}{P+R} \\
\end{aligned}
\end{equation}
The metrics are similar to those in ECPE task. Proposed causes denotes the cause clauses predicted by the model, annotated causes denotes the cause clauses that are labeled in the dataset and correct causes means the clause that are both labeled and predicted as a cause clause.

\begin{table}[h]
\centering
\caption{Details about the two ECE datasets.}
\label{ECE dataset}
\begin{tabular}{l|c|c} \hline
 & Number & Percentage \\ \hline
 \emph{Chinese Dataset} & & \\ \hline
Documents & 2105 & 100\% \\
Documents with 1 cause clause & 2046 & 97.20\% \\
Documents with 2 cause clauses & 56& 2.61\% \\
Documents with 3 cause clauses & 3 & 0.14\% \\ \hline

\emph{English Dataset} & & \\ \hline
Documents & 2145 & 100\% \\
Documents with 1 cause clause & 1949 & 90.86\% \\
Documents with 2 cause clauses & 164& 7.65\% \\
Documents with 3 cause clauses & 32 & 1.49\% \\ \hline
\end{tabular} 
\end{table}

We use pre-trained word2vec \cite{word2vec} for Chinese dataset and Glove \cite{Glove} for English dataset. Our method does not use any emotional annotations in the testing. The baselines are listed as follows:
\begin{itemize}
\item {\bf Word2vec} uses word representations obtained by Word2vec \cite{word2vec} as features and then trains an SVM classifier.
\item {\bf SVM} uses unigrams, bigrams, and trigrams as features and then trains an SVM classifier.
\item {\bf CNN} is the basic CNN proposed by \cite{CNN}. In this model, the candidate clause presentation and the emotion clause presentation are together as input.
\item {\bf Multi-kernel} uses the multi-kernel method to identify the cause  \cite{SVM-Event-DrivenEmotionCauseExtractionwithCorpusConstruction}.
\item {\bf Memnet} denotes a deep memory network \cite{AQuestionAnsweringApproachtoEmotionCauseExtraction}.
\item {\bf ConvMS-M} is a convolutional multiple-slot deep memory network \cite{AQuestionAnsweringApproachtoEmotionCauseExtraction}.
\item {\bf CANN}  denotes a co-attention neural network model \cite{ACo-AttentionNeuralNetworkModelforEmotionCauseAnalysiswithEmotionalContextAwareness}.
\item {\bf RTHN} is a joint framework, using RNN and Transformer to encode and classify multiple clauses synchronously \cite{ARNN-TransformerHierarchicalNetworkforEmotionCauseExtraction}.
\item {\bf MANN} employs a multi-attention-based model for emotion cause extraction\cite{Context-AwareMulti-ViewAttentionNetworksforEmotionCauseExtraction}.
\item {\bf CANN-E} \cite{Emotion-CausePairExtraction:ANewTasktoEmotionAnalysisinTexts} is an extended model with a removal of annotated emotions from {\bf CANN}.
\item {\bf Inter-EC} is a hierarchical model mainly proposed for ECPE task  \cite{Emotion-CausePairExtraction:ANewTasktoEmotionAnalysisinTexts}.
\end{itemize}

The results are listed in Table~\ref{ECE}. We separate these methods into two types, ECE-based and ECPE-based, with the former used emotion signals and the latter not. In the English dataset, our approach achieve the best performance with 5.6\% improvement on f1-score and 1.6\% improvement on precision compared with the ECE-based baselines. It indicates that our ECPE-based method can overcome the deficiency of emotion annotations and even perform better on ECE task. In the Chinese dataset, {\bf DQAN} outperforms 7.3\% higher on f1-score compared with the best ECPE-based method and 84\% higher compared with {\bf CANN-E}, an ECE-based model with emotion annotation removed, which exhibits that by jointly adding semantic information and weighted loss, our method is also better to identify causes in such ECE tasks. Admittedly, the performance on the Chinese dataset is lower than a few ECE-based methods. One possible explanation is that such ECE methods may overfit dataset. From the details of the two datasets, the percentage of documents with more than one cause clause is only 2.75\% in the Chinese dataset while that number is 9.09\% in the English dataset, which indicates that the Chinese dataset is a simpler one compared with the English dataset. Moreover, the distance between single cause clause and emotion clause may be much close. Thus, such ECE approaches may perform well by overfitting the distance, while they perform badly when it comes to the English dataset with more complex and diverse cause clauses.

\section{Conclusions}\label{Conclusions}
In this paper, we propose a neural network named dual-questioning attention network for emotion-cause pair extraction task. Experiment results show that our method performs better than a range of baselines on ECPE task and even better on some evaluations than other methods on ECE task, even without essential emotion annotations. Further analysis demonstrates that a proper weight setting and contextual semantics are necessary.

\bibliographystyle{IEEEtran}
\bibliography{ecpe}

\end{document}